%% file: main.tex
\documentclass[10pt,twocolumn,letterpaper]{article}

\usepackage{wacv}              

\usepackage[utf8]{inputenc} 
\usepackage[T1]{fontenc}    
\usepackage{graphicx}
\usepackage{amsfonts}       
\usepackage{booktabs}
\usepackage{stfloats}
\usepackage{xcolor}
\usepackage{latexsym}
\usepackage[fleqn]{amsmath}
\usepackage[psamsfonts]{amssymb}
\usepackage{todonotes}

\colorlet{Red}{red}
\colorlet{Yellow}{yellow}

\usepackage[pagebackref,breaklinks,colorlinks]{hyperref}

\usepackage[capitalize]{cleveref}
\crefname{section}{Sec.}{Secs.}
\Crefname{section}{Section}{Sections}
\Crefname{table}{Table}{Tables}
\crefname{table}{Tab.}{Tabs.}

\def\eg{\emph{e.g}\onedot} 

\def\ie{\emph{i.e}\onedot}

\def\wacvPaperID{660} 
\def\confName{WACV}
\def\confYear{2024}

\begin{document}

\title{VortSDF: 3D Modeling with Centroidal Voronoi Tessellation on Signed Distance Field}

\author{Diego Thomas$^1$, Briac Toussaint$^2$, Jean-Sebastien Franco$^2$, Edmond Boyer$^3$\\
$^1$Kyushu University, $^2$INRIA Grenoble Rhone-Alpes-LJK, $^3$Meta Reality Labs\\
} 

\twocolumn[{
\renewcommand\twocolumn[1][]{#1}
\maketitle
\begin{center}
  \centering
  \captionsetup{type=figure}
  \includegraphics[width=0.94\textwidth]{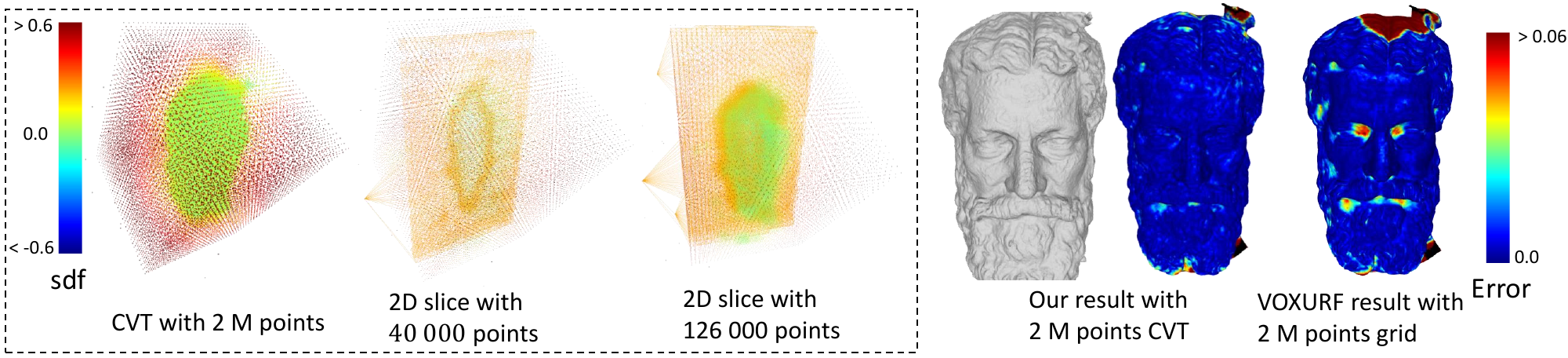}
  \caption{Our proposed method reconstructs detailed 3D surfaces using CVT that adapts to the reconstructed geometry. 
  }
  \label{fig:teaser}
\end{center}
}]

\begin{abstract}
  Volumetric shape representations have become ubiquitous in multi-view reconstruction tasks. They often build on regular voxel grids as discrete representations of 3D shape functions, such as SDF or radiance fields, either as the full shape model or as sampled instantiations of continuous representations, as with neural networks.  Despite their proven efficiency, voxel representations come with the precision versus complexity trade-off. This inherent limitation can significantly impact performance when moving away from simple and uncluttered scenes. In this paper we investigate an alternative discretization strategy with the Centroidal Voronoi Tessellation (CVT). CVTs allow to better partition the observation space with respect to  shape occupancy and to focus the discretization around shape surfaces. To leverage this discretization strategy for multi-view reconstruction, we introduce a volumetric optimization framework that combines explicit SDF fields with a shallow color network, in order to estimate 3D shape properties over tetrahedral grids. Experimental results with Chamfer statistics validate this approach with unprecedented reconstruction quality on various scenarios such as objects, open scenes or human.
\end{abstract}

\section{Introduction}
\label{sec::intro}
\input{Sections/Intro}

\section{Related works}
\label{sec::relworks}
\input{Sections/RelatedWorks}

\section{Method}
\label{sed::method}
\input{Sections/Method}

\section{Experiments}
\label{sec::experiments}
\input{Sections/Experiments}

\section{Conclusion}
We propose a novel method to reconstruct 3D geometry of a target scene from a set of multi-view images by optimizing a SDF field on a Centroidal Voronoi Tessellation. We formulate the optimization framework over the CVT and its dual tetrahedral mesh, designing an efficient framework to output detailed 3D shapes with competitive computation times. Our experimental results validate the key ideas in our proposed method and demonstrate at equivalent discretization level we can achieve a significantly higher level of extracted detail in a majority of situations, compared to competitive approaches. In a number of occurrences our method outperforms or is competitive with SOTA methods while using a magnitude lower level of discretization. Our work opens new promising directions toward detailed 3D reconstruction of large scale scenes under a contained computational time and GPU memory budget. 

\section{Acknowledgement}
\vspace{-3mm}
This work was in part supported by JSPS/KAKENHI JP23H03439 in Japan.

{\small
\bibliographystyle{ieee_fullname}
\bibliography{main}
}

\end{document}

%% file: Sections/Intro.tex

The 3D digitization of real-world objects is a foundational element for future technologies, that has motivated extensive research in recent decades. Among the primary solutions, multi-view capture systems have arisen as key tools to generate high-quality shape and appearance models of 3D scenes. Despite their effectiveness, the reconstruction of detailed geometry from multiple high-resolution images remains a challenging task due to the inherent ambiguities and complexity in the visual observations.


In multi-view 3D reconstruction, volumetric shape representations are increasingly prevalent, e.g.~\cite{mildenhall2020nerf, wang2021neus, wu2022voxurf, li2023neuralangelo}. This is in part due to their ability to relate shape properties to image observations through differential rendering. This has been extensively leveraged, particularly with networks, typically MLPs, which are trained to model shape geometry or appearance that best explain the image observations under photometric losses.


In the seminal work NeRF \cite{mildenhall2020nerf}, volumetric radiance fields are estimated by integrating color and opacity along pixel rays. While this method produces highly realistic new viewpoints, the associated geometry extracted from opacity boundaries is artifact-ridden. Subsequent works, such as NeuS \cite{wang2021neus}, address this by explicitly parametrizing the surface and its properties with signed distance fields, though they struggle to achieve the same image quality as volumetric radiance fields. Hybrid methods that combine explicit SDF grids with shallow networks for color offer benefits in both geometry and appearance modelling \cite{SunSC22, wu2022voxurf}. Yet, the majority of such methods rely on some form of regular axis-aligned grids to discretize 3D observation spaces and are therefore sensitive to the inherently poor quality-to-parsimony trade-off of these representations. Moreover, regular grids result in sub-optimal meshes when paired with the Marching Cubes algorithm \cite{lorensen1998marching} ubiquitously used for explicit mesh surface conversion. 

Voxel grids uniformly discretize the observation space, regardless of the shape’s location. Consequently, increasing resolution specifically near the shape surface requires non-trivial specializations. Octrees and HashMaps \cite{neus2, li2023neuralangelo} offer hierarchical space discretization, but their dynamic update during optimization and raymarching is cumbersome as they cannot straightforwardly follow the deformation during surface updates. Instead, we explore a hierarchical tetrahedral discretization guided by the Centroidal Voronoi Tesselation (CVT) algorithm. CVTs yield provably optimal discretizations and exhibit noteworthy advantages in our context: (i) Efficient ray marching  along rays through tetrahedra; (ii) The ability to hierarchically up-sample and deform tetrahedral grids in adaptive fashion w.r.t the encoded shape surface.

As conventional 3D convolution or automatic differentiation do not easily extend to such non-uniform cell complexes, we here develop a complete representation and methodology to encode and optimise 3D fields over them. 
Specifically, given multiple images of a scene, our approach  jointly optimizes a hierarchical CVT discretization with an associated neural field. The dual of the CVT defines a tetrahedral grid over which SDF and color feature values are stored. Images are rendered by sampling along pixel ray, with SDF and feature values interpolated at the sampled points. A color network is trained to predict the colors at sample points based on the interpolated features. We show that feature extraction and gradient back-propagation along rays can be efficiently performed over the tetrahedral dual of the CVT. Our hierarchical approach up-samples the CVT, after neural field convergence, at increasing levels of details, with a tenfold difference in grid resolution between subsequent levels. 

We target several applications representative of different reconstruction scenarios such as objects, open scenes or humans and evaluate our method with Chamfer error statistics. Our experiments on public datasets like BlendedMVS \cite{yao2020blendedmvs} and 4D Human Outfit \cite{4dhumanoutfit} demonstrate that this strategy yields significantly more reconstruction detail when compared to SOTA techniques with equivalent time and primitive budgets.  
Our main contributions are: (1) introducing CVT discretization for neural fields in multi-view reconstruction; (2) an implicit CVT optimization method that adapts to the optimized SDF field; (3) proposing an associated fast optimization framework.


%% file: Sections/RelatedWorks.tex


\begin{figure*}[t]
    \includegraphics[width=0.9\textwidth]{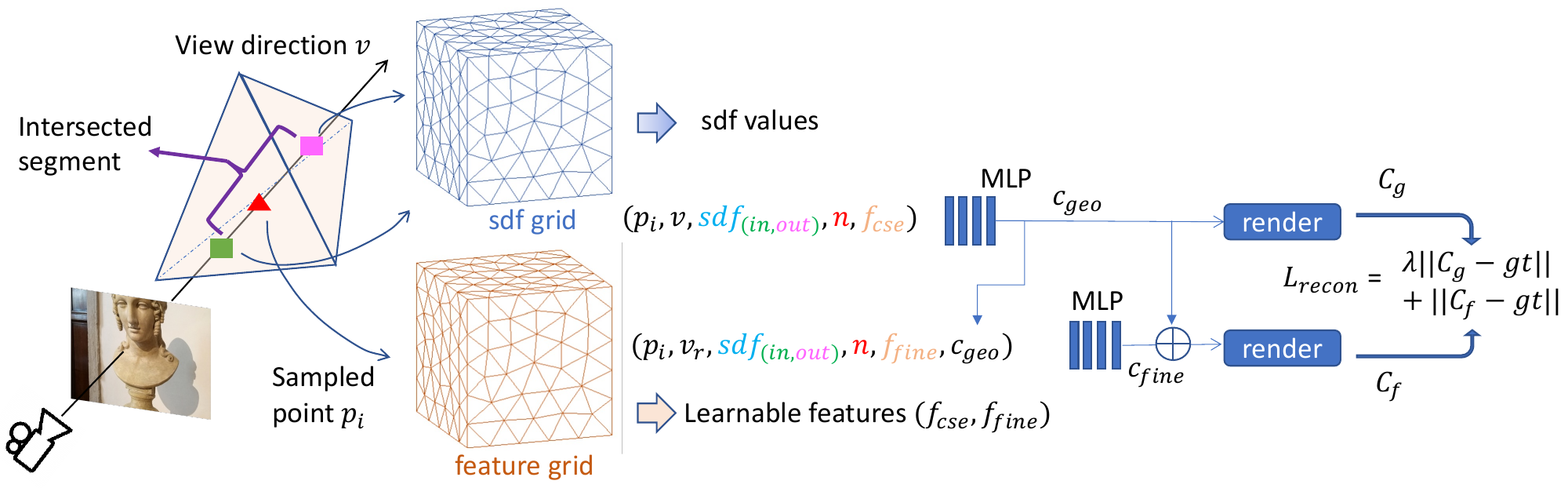}
    \centering
    \caption{We propose a volumetric optimization framework that combines explicit SDF fields and learnable features with two shallow color networks, in order to estimate 3D shape properties over tetrahedral grids. 
    }
    \vspace{-2mm}
    \label{fig::pipeline}
\end{figure*}

Neural radiance fields have been a highly active topic in recent research. Initially popularized for the problem of image-based rendering by NeRF \cite{mildenhall2020nerf}, it has been immediately explored by the community as a way to approach the multi-view 3d reconstruction problem \cite{yariv2020multiview}. NeRF relies on an implicitly parameterized neural function to approximate the plenoptic function \cite{kellnhofer2021neural}, and leverages volumetric differential rendering \cite{Lombardi2019TOG} to optimize the neural parameters. While 
providing volumetric continuity in the observation space, the neural optimization proved initially slow and inefficient for accurate and detailed surface geometry extraction, leading to several research threads of improvement.

\vspace{-1mm}
\paragraph{More detailed volumes} have been pursued by addressing limitations of neural field representations, \eg in the frequency domain with Mip-Nerf \cite{barron2021mipnerf, barron2023zipnerf}, in the bounded restriction of the spatial domain for background inclusion with Nerf++\cite{kaizhang2020}, or both~\cite{barron2022mip}. They do not however trivially transpose to higher detail surface extraction as proposed.
\paragraph{Surface-based representations} were introduced to address volumetric approaches' inherently limited ability to encode continuous shape surfaces.
While UNISURF~\cite{oechsle2021unisurf} centered its representation on occupancy to this goal, inference based on signed distance fields (SDF)~\cite{park2019deepsdf} proved more successful, as initially explored by IDR \cite{yariv2020multiview}, and later improved by Neus~\cite{wang2021neus} by modeling shape uncertainty through a local volume integration around the main surface mode with a Gaussian opacity profile, which remains competitive to this day and serves as a key benchmark.  
\vspace{-3mm}
\paragraph{Computational efficiency} has been an essential topic to bring Nerf and derivatives to a more usable realm than the initial 30+ hour optimization time. Parting with the implicit network representation in favor of an explicit uniform volumetric grid, either with a fully explicit non-neural radiance parametrization~\cite{fridovich2022plenoxels}, or using shallow networks parametrized by grid features embedded in the volume~\cite{reiser2021kilonerf,sun2022direct} or onto coordinate-projected hyperplanes for parameter dimensionality reduction~\cite{chen2022tensorf,fridovich2023k,chen2023factor}, have been central ideas leading to computational time improvements of several orders of magnitude. Of particular interest to us are the attempts to spatially sparsify and hierarchize the volume to focus resources on shape boundaries, using \eg an octree structure \cite{liu2020neural,yu2021plenoctrees}, and simultaneously improve representation performance through multiscale features~\cite{mueller2022instant}. These order of magnitude improvements to volumetric methods were recently transposed to the realm of surface methods with predictive performance that is on par~\cite{neus2} or improved \cite{wu2022voxurf}. Recent methods leverage the representational advantage for surface extraction of extended scenes~\cite{li2023neuralangelo}. Yet, all these representations are fundamentally tied to an axis aligned grid core, which we show inhibits access to even higher surface estimation performance.


\vspace{-1mm}
\paragraph{Adaptive non-uniform spatial discretization.}
A number of methods have explored non-uniform cell sampling or decomposition of space as a way to optimally sample and focus resources on the surface vicinity. The recent splatting approaches ~\cite{kerbl20233d, chen2023neurbf} for example opt for blob primitives whose extent and positions are optimized jointly with adjoining opacities and colors, to explain input views. Of particular relevance to our work, DeRF~\cite{rebain2021derf} optimizes a Voronoi cell decomposition to improve overall performance in a volumetric setting, but does not explore coarse-to-fine and surface adaptive sampling as we propose. TetraNerf~\cite{kulhanek2023tetra} uses a Delaunay marching structure to slightly accelerate volumetric radiance field inference down to 10-20 hours, but the decomposition is fixed, based on a pre-computed set of COLMAP points. None of these approaches deal with surface extraction and leverage the cell structure to efficiently perform this task, as proposed. 

To provide this key improvement and still benefit from the associated computational performance boost, we note that two surface extraction algorithms based on CVTs \cite{wang2016volumetric} were shown to significantly outperform  Marching Cubes and Delaunay Tessellation for exactly this task. The first such variant clips the cells by looking at intersection between edges and implicit surface; and the second one refines the tessellation by adding vertices at the intersection between surface and bisector of edges. The work demonstrates CVTs to be a regular tessellation guaranteed to be manifold, with excellent surface accuracy achieved with a predefined number of cells. It is also notable that efficient ray-marching algorithms exist for cell-based 3D structures~\cite{aman2022compact}. Our novel framework builds on these properties to propose optimally adapted sampling of 3D space around the surface of interest, for the purpose of neural adaptive 3D surface reconstruction.

%% file: Sections/Method.tex
Given a set of multiple images, our method jointly optimizes an SDF field, discretized on a hierarchical CVT, and two view-dependent shallow color networks to predict view-dependent color at any 3D location, inspired by recent works on direct SDF optimization with uniform voxel grids~\cite{sun2022direct, wu2022voxurf}. Those color networks are queried at sampled points along pixel rays defined by the camera location and random pixel coordinates. 

As illustrated in the method outline (Figure~\ref{fig::pipeline}), the first color network takes as input a 3D location, a 3D direction vector, SDF values,  a normal vector and  8-dimensional learnable features $f_{cse}$. The second network is a color refinement network that takes as additional input the coarse color and uses an additional 8-dimensional learnable feature set $f_{fine}$. Note that the refinement network takes the reflected vector $v_r$ instead of the viewing direction $v$ in order for the network to focus more on specular effects. 

SDF and feature values are optimized by back-propagating a photometric loss from the sampled points to the CVT sites. After convergence, the CVT is iteratively refined and up-sampled non-uniformly with respect to the current estimate of the SDF field, therefore increasing the shape resolution at the vicinity of the shape surface. 

\subsection{Centroidal Voronoi Tesselation}
A tessellation of a 3D space is a disjoint set of polyhedron that fills the 3D space of interest. 
CVTs are used in a wide range of applications, in Visual Computing and beyond, as it provides an elegant tool to compute a regular and optimized discretization of the 3D space \cite{wang:hal-01185210}. Given a set of points, called sites, a Voronoi tessellation partitions the space into regions around the sites and is dual to the Delaunay triangulation of these sites. 

A Voronoi cell $V_i$ is associated to its site $x_i$ and is composed of all points  that are closest to $x_i$: \\ 
\begin{equation*}
    \{ p \in \ R^3 \ / \ ||p-x_i|| < ||p-x_j||, \ j \in [1,K], j\ \neq\ i\}
\end{equation*}

Voronoi cells are delimited by segments in 2D and convex polygons in 3D that are the intersection of the bisectors between pairs of sites. When the sites coincide with the centroids of the associated cells, the tessellation is called a Centroidal Voronoi Tessellation. Intuitively, CVT cells optimally partition the input domain as k-means clusters minimizing a variance or quantification error~\cite{du99}.


\vspace{-1mm}
\subsection{Coarse-to-fine Centroidal Voronoi Tesselation}
Central to our proposed method is densifying the continuous Signed Distance Function (SDF) discretization near the surface. Unlike previous work such as DMTet~\cite{shen2021dmtet}, we do not know the true SDF. Instead, we simultaneously optimize both the SDF values and their discretization, which presents a significantly greater challenge. Given a 3D region of interest, such as a bounding box, we begin with a coarse uniform grid, for example, $16 \times 16 \times 16$ in our experiments. From this grid, a coarse initial Centroidal Voronoi Tessellation (CVT) is generated by minimizing the CVT energy that moves the sites to the center of their cell. Additionally, we add the center of each camera as complementary sites to the CVT in order to speed-up ray-marching operations, as will be elaborated on in Section~\ref{sec::raymarching}.


After the optimization of the color network parameters and of the SDF and feature values within the CVT, we up-sample the discretization by adding a point at the center of each surface-crossing edge of the dual of the CVT, \ie of the associated  tetrahedral Delaunay mesh. Edges that present a SDF value smaller than $1.5$ the edge length at one end point are also up-sampled. The CVT energy is then minimized, at each up-sampling iteration, to ensure that the vertices are locally uniformly distributed, which is crucial to reduce sampling artifacts. This process is repeated until the expected level of details is reached.

We keep a KD-tree for each level of discretization and compute the K-nearest neighbors of each site 
to enable propagation of gradients and smoothing. 

\begin{figure}[t]
    \includegraphics[width=0.45\textwidth]{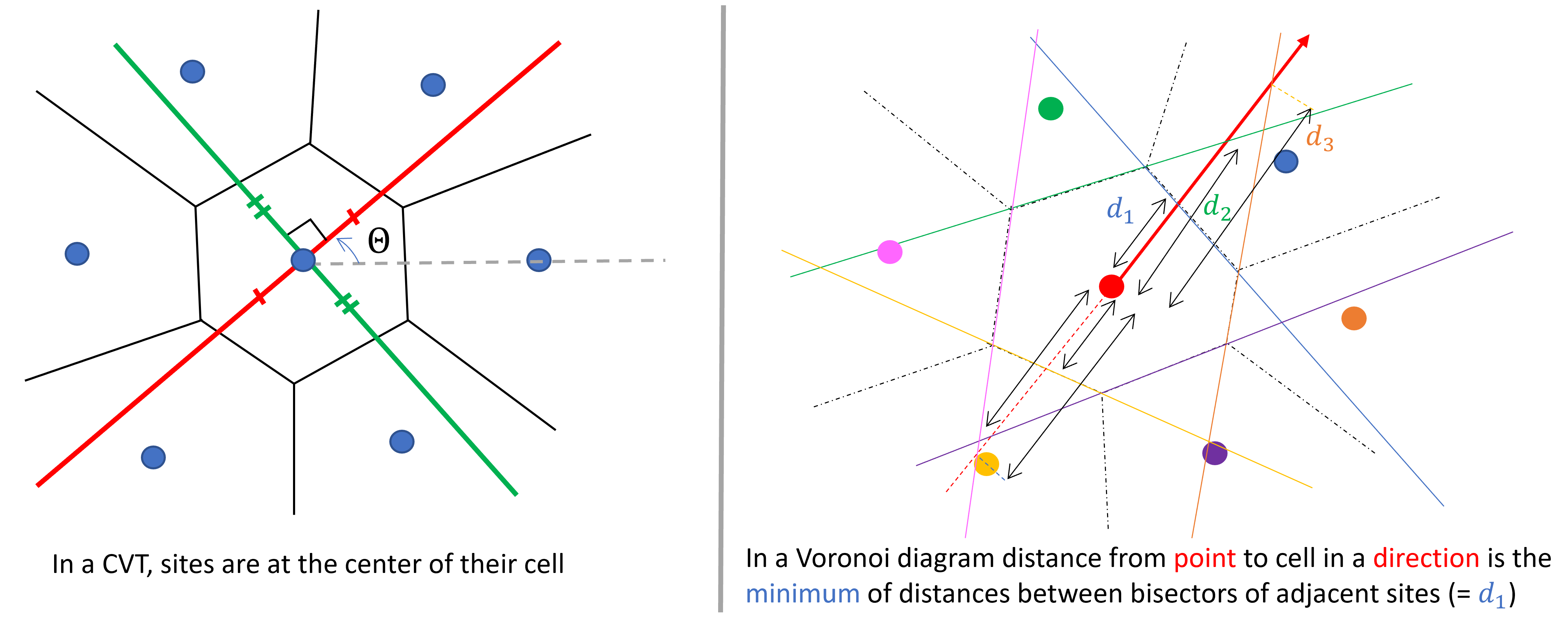}
    \centering
    \caption{The site locations are optimized using an approximated CVT algorithm that does not explicitly identify the Voronoi cells (left) but consider  the neighboring bisectors instead (right).}
    \vspace{-2mm}
    \label{fig::CVTIMP}
\end{figure}

\vspace{-1mm}
\subsubsection{SDF-aware implicit CVT optimization}
High quality 3D reconstructions require very fine discretization, in practice millions of sites. At such scale, traditional CVT optimizations, even those based on the  L-BFGS quasi-Newton method~\cite{liu2009centroidal}, become computationally prohibitive,  since the explicit boundaries of each cell must be recomputed every time the sites are moved. While full GPU solutions have been proposed \cite{ray2018meshless} they still struggle with millions of sites. In addition, we expect the CVT to adapt locally to the optimized SDF field so that the shape surface is materialized by the Voronoi cell faces. Inspired by \cite{maruani2023voromesh}, we propose to build an approximate CVT, with significant computational benefits yet providing nearly equivalent behaviour in cell spacing. With millions of sites, computing the exact Voronoi diagram with off-the-shelf libraries takes about $5$ minutes. Then $30$ iterations of the CVT optimization requires about $2.5$ hours. With our approximated CVT, one iteration takes about $1$ second and we can run $300$ iterations in about $5$ minutes.


As mentioned earlier, a CVT is obtained when its sites are the centroids of the associated Voronoi diagram.  Traditional algorithms iterate therefore between estimating the Voronoi diagram and moving the sites towards the cell centroids. With the aim to avoid the explicit estimation of the Voronoi diagram when optimizing the site positions, we observe that a CVT site should be equidistant from the border of its Voronoi cells in any direction around the site (see fig. \ref{fig::CVTIMP}). Such Voronoi cells are spanned by the bisectors between sites. Hence we propose to define  our CVT loss using differences between distances to the neighbouring bisectors, instead of distances to identified Voronoi cells. To this purpose we randomly sample directions around a site using two angles $(\Theta, \Phi)$  and sum the distance differences to the closest bisectors along each direction.  In practice we use $N = 24$  nearest neighbours\footnote{We re-estimate the K-nearest neighbors every $100$ iterations.} around a site to estimate the bisectors and sample in $3$ orthogonal directions obtained by rotating the cartesian basis with the random angles $(\Theta, \Phi)$. Note that these  angles are different for each site and change at each iteration. The CVT loss writes then
\begin{equation}
    L_{CVT} = \frac{1}{2} \sum_{s_i} \sum_{j=0,1,2} (d(\mathbf{s_i}, \mathbf{e_j}(\Theta, \Phi)) - d(\mathbf{s_i}, -\mathbf{e_j}(\Theta, \Phi)))^2.
\end{equation}
$\left\{\mathbf{e_0}(\Theta, \Phi), \mathbf{e_1}(\Theta, \Phi), \mathbf{e_2}(\Theta, \Phi)\right\}$ is the rotated cartesian basis and $d()$ is defined as
\begin{equation}
    d(\mathbf{s_i}, \mathbf{r}) = min_{\mathbf{s_j} \in N(\mathbf{s_i})}(\|\mathbf{s_i} - b(\mathbf{s_i}, \mathbf{s_j}, \mathbf{r})\|_2),
    \label{eq::CVT distance}
\end{equation}
where $b(\mathbf{s_i}, \mathbf{s_j}, \mathbf{r})$ \textcolor{blue}{is} the distance from $\mathbf{s_i}$  to its bisector with $\mathbf{s_j}$ in the direction $\mathbf{r}$. 

In addition, in case when the SDF values of $\mathbf{s_i}$ and $\mathbf{s_j}$ have opposite signs we move the bisector plane so that it lies on the $0$ crossing. The intuition is that the bisectors of the CVT cells coincide with the middle of the dual tetrahedra. Thus the points that are sampled at the middle of the segments that intersect ray and tetrahedron will lie closer to the $0$ level set. As a consequence the color networks would be optimized closer to the surface. 

We implement the losses on the GPU, compute the explicit gradients and use PyTorch Adam optimizer to minimise $L_{CVT}$ with $300$ iterations. After the CVT is optimized, the Delaunay tetrahedra are computed to restart the SDF and features optimization.

\begin{figure}[t]    \includegraphics[width=0.45\textwidth]{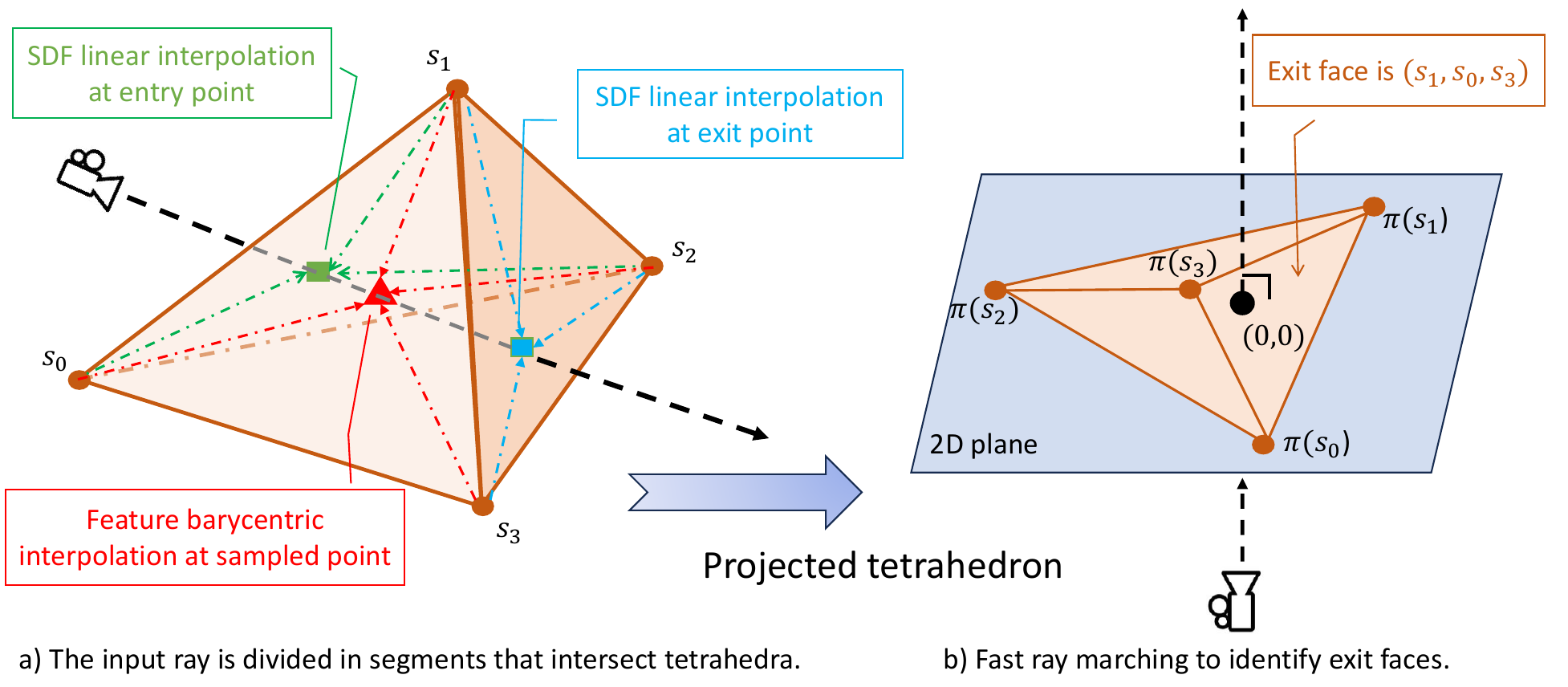}
    \centering
    \caption{a) The SDF values at the intersecting segments extremities are obtained by linear interpolation of the SDF values at the entry and exit faces. Features at sampled points are linearly interpolated from the $4$ vertices of the tetrahedron. 
    b) The viewing ray exit face is the face to which camera center belongs after projecting the tetrahedron into the plane perpendicular to the viewing ray.
    }
    \vspace{-2mm}
    \label{fig::features}
\end{figure}



%
%

\vspace{-1mm}
\subsection{Differentiable rendering in a tetrahedral mesh}
\label{sec::raymarching}
Differentiable rendering is a key component of volumetric reconstruction strategies. In such rendering, efficient sampling of points along pixel viewing lines is essential. However, doing efficient sampling on non-uniform tetrahedral grids requires specific algorithms.  


We adapt the $32$ bits tetrahedral structure proposed in~\cite{aman2022compact} for fast ray-marching. 
We walk through the tetrahedral mesh and along a viewing line by: (i) finding the exit face in the current tetrahedron; (ii) identifying the next tetrahedron. 
To find the exit face we project all vertices in the plane centered at the camera center and which normal equals  the ray direction vector. Then the exit face is the face that contains the origin in the projected coordinate system (the entry face is not counted). See figure \ref{fig::features} for illustration of the process. 

In contrast to \cite{aman2022compact} and  \cite{kulhanek2023tetra}, we include the camera centers in the tetrahedral mesh. This way, the entry point of a ray in the tetrahedral mesh is easily obtained as we only need to check the tetrahedron that contains the camera center. The output of our ray marching algorithm is a list of visited tetrahedra in association with their entry and exit points. To speed up computations, we prune out all the tetrahedra that exhibit a null contribution to the rendered image.

\vspace{-1mm}
\subsubsection{Volumetric rendering}
The 3D ray $(\mathbf{o}, \mathbf{v})$ originates from the camera center $\mathbf{o}$ in the direction $\mathbf{v}$. When intersecting the tetrahedral grid, it is split into $n$ segments $\{s(t) = [\mathbf{in}(t): \mathbf{out}(t)]| 0 \leq t \leq n \}$ with non-null contribution to the accumulated color using the output of our proposed ray-marching algorithm\footnote{We use a maximum of $n=1024$ segments per ray in our experiments.}. For each segment we sample a point $\mathbf{p}(t) = \frac{\mathbf{in}(t)+\mathbf{out}(t)}{2} $ at the middle of the segment and query the color networks $g_{geo}$ and $g_{fine}$ at each point $\mathbf{p}(t)$. The color of the ray is then: 
%
\begin{flalign}
    \mathbf{C(o, v)} = \sum_{t=0}^{n}{\omega(t)c(\mathbf{p}(t), \mathbf{v})}, \\ 
    \label{eq::render}
    \omega(t) = \alpha(t)T(t), \\ 
    T(t) = \prod_{i < t}(1-\alpha_i),
\end{flalign}
\vspace{-1mm}
where $\mathbf{C(o, v)}$ is the estimated color for this ray, $\omega(t)$ a weight for the point $\mathbf{p}(t)$, and $\mathbf{c(p(t), v)}$ the color at the point $\mathbf{p}$ along the viewing direction $\mathbf{v}$ that is the output of either $g_{geo}$ or $g_{fine}$. $\alpha(t)$ is the transmittance at $t^{th}$ point and $T(t)$ is the accumulated transmittance. Different strategies exist to compute the transmittance,  a standard one being to use a volume rendering formulation \cite{mildenhall2020nerf}. A more efficient strategy is to use the normalized S-density as weights~\cite{wang2021neus}: 
\begin{flalign}
    \alpha_t = clip\left(\frac{1+e^{-\beta sdf(\mathbf{in}(t))}}{1+e^{-\beta sdf(\mathbf{out}(t)}},0,1\right),
\end{flalign}
where $sdf(\mathbf{in}(t))$ is the SDF at the entry point in the $t^{th}$ segment and $\beta$ is a scale factor that is gradually increased during optimization. The $clip$ function clamps the transmittance value between $0$ and $1$. When $\beta$ becomes large, long segments may generate too small gradients as the differences in SDF values between the entry and exit points become too large. Consequently, we further subdivide segments that cross the surface. 

The SDF values for each entry and exit points are obtained by linearly interpolating the SDF values at the three vertices of the entry and exit faces (respectively). These two SDF values form the geometric feature $f_{geo}$ of the coarse color network. Similarly we compute the normal vectors of the entry and exit points and add these normal vectors as input of the refinement network.

We render colors for both the coarse and fine color networks and obtain the corresponding colors $\mathbf{C_i^{geo}}$ and $\mathbf{C_i^{fine}}$, respectively, at pixel $i$. Given the ground truth color $\mathbf{C_i}$ at this pixel,  we get the following photometric data term for the SDF optimization:
%
\begin{equation*}
    E_{rgb} = \sum_{i \in [1:N]} \frac{\lambda\|(\mathbf{C_i}-\mathbf{C_i^{geo}})\|_2^2 + \|(\mathbf{C_i}-\mathbf{C_i^{fine}})\|_2^2}{(\|\mathbf{C_i}\|_2 + \epsilon)},
\end{equation*}
where $\lambda$ is a weight ($1$ at the coarser stage then $0.5$) and $\epsilon$ is a small value ($0.1$ in our experiments). 
 
\subsection{SDF field regularization}
While fully implicit surface representations are naturally regularized by the weights of the neural network, special attention is required to regularize discrete SDF fields. 

\vspace{-1mm}
\subsubsection{Normal Smoothing}
Since SDF values are linearly interpolated within a tetrahedron, we can express the gradient of the SDF function within a tetrahedron as a function of the values at the tetrahedron's vertices and of the spatial gradients $\mathbf{\nabla w_i}$ of the interpolation weights $w_i$. By solving a linear system in each tetrahedron we compute the gradient vector associated to each tetrahedron.  We can then express the spatial gradient of the SDF field inside each tetrahedron as a function of the SDF values at the four summits.
\begin{flalign}
    \mathbf{\nabla sdf_t} = \sum_i \mathbf{\nabla w_i}sdf(i). 
\end{flalign}
Each gradient $\mathbf{\nabla sdf_t}$ linearly depends on the SDF values at the $4$ vertex of a tetrahedron.

We use a smoothing regulator that aligns the gradients of the sdf and the gradients of the smoothed sdf values. 
%

\vspace{-3mm}
\begin{flalign*}
    L_{reg} = 0.5\sum_t (1-(\frac{\mathbf{\nabla sdf_t} \cdot \mathbf{\nabla sdf_t}^{smooth}}{\|\mathbf{\nabla sdf_t}\|_2\|\mathbf{\nabla sdf_t}^{smooth}\|_2})^2).
\end{flalign*}
\vspace{-5mm}
\subsubsection{Smoothing with K nearest neighbors.} One key difficulty in using tetrahedral grid is that given a 3D point in space we cannot directly access the tetrahedra that contains the point. As a consequence it is not possible to average SDF values sampled uniformly around a summit of the tetrahedral grid. In addition, the tetrahedral grid is non uniform so simply averaging SDF values at summits of adjacent tetrahedra creates unwanted artifacts. Therefore, we compute the smoothed SDF values on the CVT by using weighted average of SDF values of K-nearest sites in the current CVT. 
Note that the K-nearest neighbors are computed only once at each up-sampling step using the corresponding KD-trees.

We also use the total variation loss defined on the edges of the tetrahedral mesh.
The final SDF gradient writes:

\begin{flalign*}
\frac{\partial sdf}{\partial t} = \frac{\partial   E_{rgb}}{\partial t} + 
w_{reg}\frac{\partial L_{reg}}{\partial t} +
w_{tv}\frac{\partial L_{TV}}{\partial t},
\end{flalign*}
where $w_{reg}$ and $w_{tv}$ are weight factors and $L_{TV}$ is a total variation loss.


%% file: Sections/Experiments.tex
\begin{figure*}[t]
    \includegraphics[width=\textwidth]{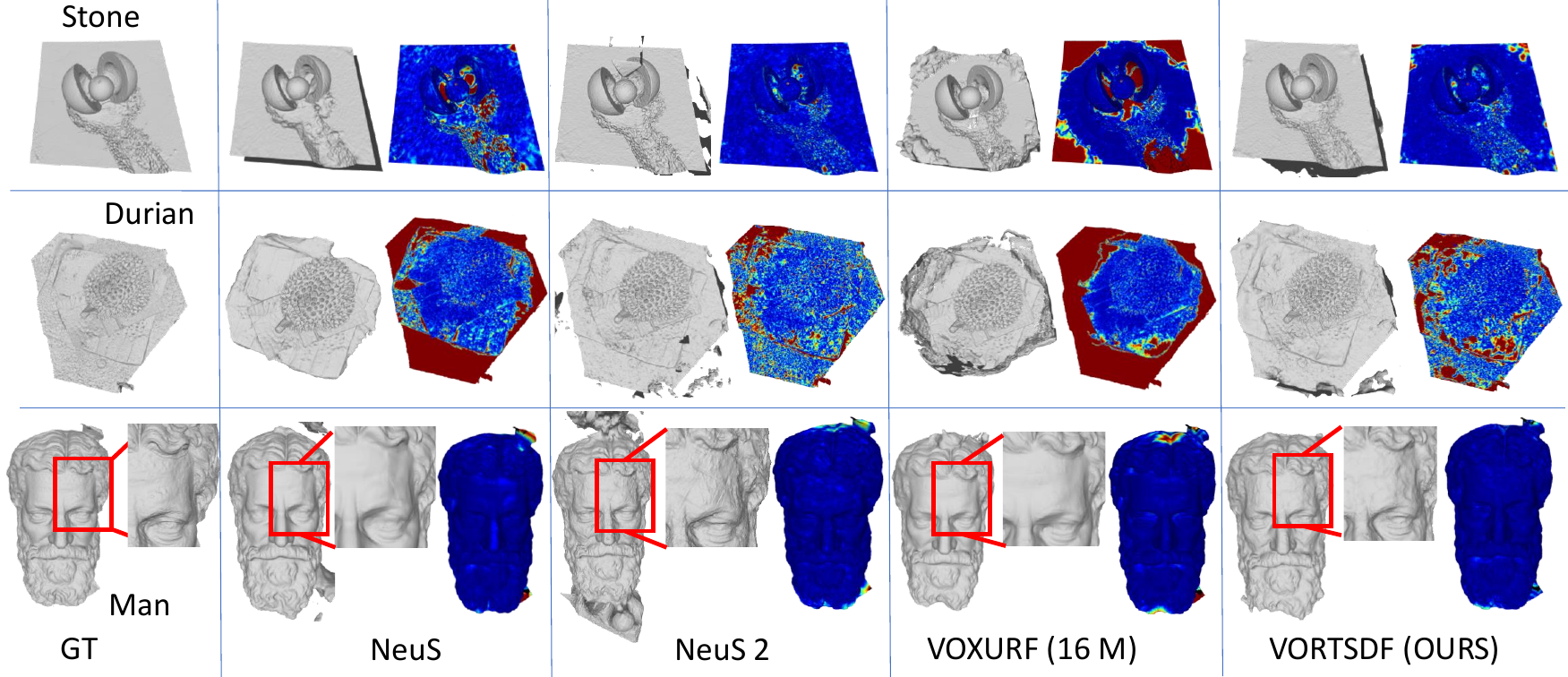}
    \caption{Comparative results we obtained with our method, NeuS and Voxurf on data "Stone", "Durian" and "Man" of BlendedMVS. We output the final 3D meshes using Marching Cubes (MC) for NeuS and Voxurf and Marching Tetrahedra for our method. We also show errors from ground truth meshes to predicted meshes as heatmaps.}
    \label{fig::results}
\end{figure*}


We evaluate the ability of our approach to reconstruct detailed 3D surfaces compared with the state-of-the-art methods NeuS \cite{wang2021neus}, Neus2~\cite{neus2} and Voxurf \cite{wu2022voxurf}. 
We use the code provided by the authors for both NeuS, Neus2 and Voxurf with recommended parameters and run our experiments on an RTX3090 GPU. 
We qualitatively and quantitatively evaluate our method on a subset of the BlendedMVS dataset \cite{yao2020blendedmvs} and a subset of the 4D Human Outfit dataset \cite{4dhumanoutfit}. 

\subsection{Metrics}
We evaluate the quality of the estimated geometry using the available ground truth 3D meshes and a point to mesh Chamfer distance. We compute these errors from the ground truth mesh to the predicted meshes to obtain accuracy $Acc$\footnote{When computing $Acc$, cluttered regions are naturally removed.} and from the predicted mesh to the ground truth mesh to obtain completeness $Compl$. We clip the errors to a maximum distance of $0.1$ meter. 
Note that smaller values are better for these metrics.


%

\subsection{Experiments on BlendedMVS dataset}
We used $7$ scenes of the blendedMVS dataset, to evaluate the ability of our method to reconstruct detailed geometry by comparing to the state of the art. Two different scenarios occur: uniform scenes where objects occupy most of the bounding box, which favors in principle uniform discretizations. Second, more open scenes with significant amount of empty space in the bounding box (Stone or Durian) to confirm the advantage of our adaptive discretization. 

We evaluate our method at the last $2$ levels of densification. Our method usually terminates with about $2$M points at lvl $5$ and about $500$K points at lvl $4$\footnote{In our experiments we performed up-sampling every $10000$ iterations.}. In comparison VOXURF uses a uniform grid of $256\times 256\times 256$ voxels, which corresponds to about $16$M points. 

\begin{table*}[t]
   \centering
    \caption{Average geometric accuracy $Acc$ (mm) (lower is better) and completeness $compl$(mm) (lower is better) obtained with our  method, NeuS2 and Voxurf, for each of the $7$ test scenes. We highlight the \colorbox{Red!50}{\textbf{best}} and \colorbox{Yellow!50}{\underline{second}} values.}
    \begin{tabular}{|c||c|c||c|c||c|c||c|c|}
    \hline
     & \multicolumn{2}{|c||}{NeuS} & \multicolumn{2}{|c||}{NeuS2} & \multicolumn{2}{|c||}{Voxurf } & \multicolumn{2}{|c|}{Ours (lvl $4$/ lvl $5$))}  \\
    \hline
     & Acc $\downarrow$ & Compl $\downarrow$ & Acc $\downarrow$ & Compl $\downarrow$ & Acc $\downarrow$ & Compl $\downarrow$ & Acc $\downarrow$ & Compl $\downarrow$  \\
    \hline
     Dog ($31$ images) & $1.54$ & $19.1$ & $1.79$ & \colorbox{Red!50}{$\mathbf{4.81}$} & \colorbox{Yellow!50}{$\underline{0.98}$} & $11.6$ & $1.83/$\colorbox{Red!50}{$\mathbf{0.96}$} & $18.72/$\colorbox{Yellow!50}{$\underline{9.88}$} \\
     \hline
     Bear ($123$ images) & $6.10$ & \colorbox{Red!50}{$\mathbf{11.25}$} & \colorbox{Yellow!50}{$\underline{2.07}$} & $70.8$ & $3.17$ & $44.08$ & $2.20/$\colorbox{Red!50}{$\mathbf{1.41}$} & $33.3/$\colorbox{Yellow!50}{$\underline{26.3}$} \\
     \hline
     Clock ($143$ images)& $1.34$ & $11.8$ & $1.11$ & $3.29$ & \colorbox{Yellow!50}{$\underline{0.95}$} & \colorbox{Red!50}{$\mathbf{0.82}$} & $0.99/$\colorbox{Red!50}{$\mathbf{0.79}$} & $1.67/$\colorbox{Yellow!50}{$\underline{1.35}$} \\
     \hline
     Durian ($124$ images)& $23.85$ & $44.3$ & \colorbox{Yellow!50}{$\underline{12.48}$} & $70.2$ & $23.34$ & $63.7$ & $17.6/$\colorbox{Red!50}{$\mathbf{11.43}$} & \colorbox{Yellow!50}{$\underline{42.8}$}$/$\colorbox{Red!50}{$\mathbf{41.1}$}  \\
     \hline
     Man ($24$ images)& $2.52$ & $49.0$ & \colorbox{Red!50}{$\mathbf{2.18}$} & $50.5$ & $2.89$ & $34.33$ & $2.51/$\colorbox{Yellow!50}{$\underline{2.31}$} & \colorbox{Yellow!50}{$\underline{27.2}$}$/$\colorbox{Red!50}{$\mathbf{19.10}$}  \\
     \hline
     Sculpt ($79$ images)& $1.95$ & $10.0$ & $1.45$ & $4.54$ & 1.34 & \colorbox{Yellow!50}{$\underline{3.63}$} & \colorbox{Yellow!50}{$\underline{1.14}$}$/$\colorbox{Red!50}{$\mathbf{1.03}$} & $4.41/$\colorbox{Red!50}{$\mathbf{3.40}$} \\
     \hline
     Stone ($56$ images)& $10.30$ & $43.28$ & \colorbox{Yellow!50}{$\underline{4.35}$} & $46.99$ & $14.53$ & $44.26$ & $8.48/$\colorbox{Red!50}{$\mathbf{3.93}$} & \colorbox{Yellow!50}{$\underline{27.10}$}$/$\colorbox{Red!50}{$\mathbf{24.73}$} \\
          
   \hline
    \hline
     Avg & $6.31$ & $26.37$ & \colorbox{Yellow!50}{$\underline{3.75}$} & $31.38$ & $6.79$ & $26.54$ & $4.96/$\colorbox{Red!50}{$\mathbf{3.30}$} & \colorbox{Yellow!50}{$\underline{22.17}$}$/$\colorbox{Red!50}{$\mathbf{20.57}$} \\
     \hline
     Time & \multicolumn{2}{|c||}{$8$h$30$mn}& \multicolumn{2}{|c||}{$5$mn} & \multicolumn{2}{|c||}{$45$mn} & \multicolumn{2}{|c|}{$30$mn /$50$mn} \\

    \hline
    \end{tabular}
    \label{tab:chamferIoU}
\end{table*}

Table \ref{tab:chamferIoU} shows the quantitative evaluation compared to NeUS, NeuS2 and Voxurf. Our method almost always improves the accuracy or ranks second to best of the reconstructed 3D mesh compared to these SOTA methods. As expected, we observe more significant gains against VOXURF with large scenes for which uniform space discretizations are not well suited. This confirms that using an adaptive CVT to support the SDF field optimization is an effective solution that yields higher frequency details in the reconstructed geometry and appearance. 

Our experimental results also show that, our method retrieves accurate reconstructed meshes already at lvl $4$, which only contains about $500$K points ($30$ times less than VOXURF). Our method even obtains better results than VOXURF at lvl $4$ for the data Sculpt. Our method is also able to produce significantly fewer artifacts than other methods as shown by the completeness results. Our method converges in about $30$mn at level $4$ and about $50$mn at level $5$. In comparison, NeuS converges in about $8$h$30$mn, NeuS2 in $5$mn and VOXURF in $45$mn.

Figure \ref{fig::results} shows qualitative comparisons. They demonstrate that denser discretizations around the surface can effectively yield higher frequency details and less outliers than other methods. Note in particular that Neus2 has some discretization artifacts in the reconstructed meshes.




\begin{figure*}[t]
    \includegraphics[width=\textwidth]{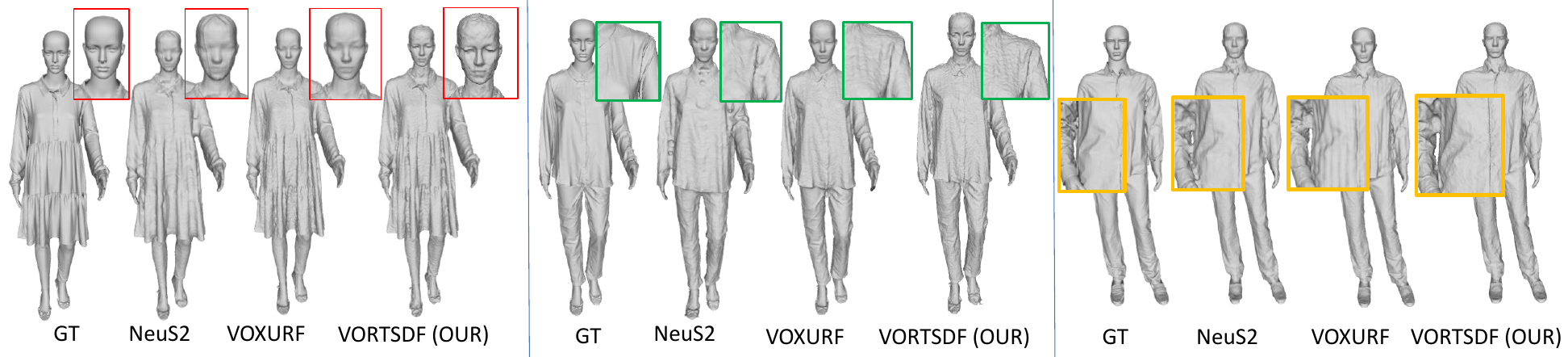}
    \caption{Comparative results we obtained with our method, NeuS and Voxurf on the 4D Human Outfit dataset. We output the final 3D meshes using Marching Cubes (MC) for NeuS and Voxurf and Marching Tetrahedra for our method.}
    \vspace{-2mm}
    \label{fig::results_m}
\end{figure*}

\subsection{Experiments on 4D Human Outfit dataset}
We used $8$ scenes of the 4D Human Outfit dataset. This dataset contains sets of $63$ high resolution images with calibrated cameras and ground truth 3D mesh captured with a millimeter precision laser scanner. With this dataset we evaluate the ability of VortSDF to reconstruct detailed geometry such as clothing wrinkles. Note that the mannequins have arms close to the body, which makes it quite favorable to uniform discretization methods. 
Yet Table \ref{tab:chamferMannekin} shows that VortSDF obtained better results in most scenes. VortSDF always obtained better results than NeuS2. The advantage VortSDF is more evident on Figure \ref{fig::results_m} where we can see significantly better level of details in the face and wrinkles. 

\begin{table*}[t]
   \centering
    \caption{Average geometric accuracy $Acc$ (mm) (lower is better) and completeness $Compl$ (mm) (lower is better) obtained with our  method, NeuS2 and Voxurf, for each of the $3$ test scenes of 4D Human Outfit dataset. We highlight the \colorbox{Red!50}{\textbf{best}} and \colorbox{Yellow!50}{\underline{second}} values.}
    \begin{tabular}{|c||c|c||c|c||c|c|}
    \hline
     & \multicolumn{2}{|c||}{NeuS2} & \multicolumn{2}{|c||}{Voxurf} & \multicolumn{2}{|c|}{Ours (lvl $3$/ lvl $4$/ lvl $5$)}  \\
    \hline
     & Acc $\downarrow$ & Compl $\downarrow$ & Acc $\downarrow$ & Compl $\downarrow$ & Acc $\downarrow$ & Compl $\downarrow$  \\
    \hline
     f-cos-hx & $4.64$ & $3.17$  & $1.76$ & $2.32$ & $2.54/$\colorbox{Yellow!50}{$\underline{1.65}$}$/$\colorbox{Red!50}{$\mathbf{1.56}$} & $2.68/$\colorbox{Yellow!50}{$\underline{1.99}$}$/$\colorbox{Red!50}{$\mathbf{1.91}$} \\
     \hline
     f-jea-hx & $5.35$ & $4.89$ & $2.25$ & $2.54$ & $2.20/$\colorbox{Red!50}{$\mathbf{1.73}$}$/$\colorbox{Yellow!50}{$\underline{1.78}$} & $2.25/$\colorbox{Red!50}{$\mathbf{1.91}$}$/$\colorbox{Yellow!50}{$\underline{1.94}$}    \\
     \hline
     f-opt1-hx & $2.34$ & $2.25$ & \colorbox{Red!50}{$\mathbf{1.57}$}  & \colorbox{Red!50}{$\mathbf{2.09}$} & $1.85/1.70/$\colorbox{Yellow!50}{$\underline{1.69}$} & $2.91/2.33/$\colorbox{Yellow!50}{$\underline{2.24}$} \\
     \hline
     f-opt2-hx & $3.23$ & \colorbox{Yellow!50}{$\underline{2.77}$} & \colorbox{Yellow!50}{$\underline{1.87}$} & \colorbox{Red!50}{$\mathbf{2.33}$} & $2.36/1.89/$\colorbox{Red!50}{$\mathbf{1.84}$} & $3.27/2.87/2.79$  \\
     \hline
     f-opt3-hx & $2.27$ & $2.21$ & $1.39$ & \colorbox{Yellow!50}{$\underline{1.91}$} & $1.51/$\colorbox{Yellow!50}{$\underline{1.33}$}$/$\colorbox{Red!50}{$\mathbf{1.29}$} & $2.14/1.93/$\colorbox{Red!50}{$\mathbf{1.86}$}  \\
     \hline
     f-sho-hx & $2.26$ & $2.11$ & \colorbox{Red!50}{$\mathbf{1.38}$} & $1.84$ & $1.60/1.45/$\colorbox{Yellow!50}{$\underline{1.39}$} & $2.03/$\colorbox{Yellow!50}{$\underline{1.77}$}$/$\colorbox{Red!50}{$\mathbf{1.73}$} \\
     \hline
                
     m-jea-hx & $1.79$ & $1.54$ & $1.27$ & \colorbox{Yellow!50}{$\underline{1.32}$} & $1.25/$\colorbox{Yellow!50}{$\underline{1.22}$}$/$\colorbox{Red!50}{$\mathbf{1.10}$} & $2.16/2.09/$\colorbox{Red!50}{$\mathbf{1.09}$}  \\
     \hline
     m-opt-hx & $2.89$ & $2.37$ & $3.35$ & $2.94$ & $2.35/$\colorbox{Yellow!50}{$\underline{2.32}$}$/$\colorbox{Red!50}{$\mathbf{1.89}$} & $2.49/$\colorbox{Yellow!50}{$\underline{2.35}$}$/$\colorbox{Red!50}{$\mathbf{2.15}$} \\
     \hline
          
    \hline
    \hline
     Avg & $3.09$ & $2.66$ & $1.68$ & $2.16$ & $1.96/$\colorbox{Yellow!50}{$\underline{1.66}$}$/$\colorbox{Red!50}{$\mathbf{1.57}$} & $2.49/$\colorbox{Yellow!50}{$\underline{2.15}$}$/$\colorbox{Red!50}{$\mathbf{1.96}$} \\
     \hline

    \hline
    \end{tabular}
    \label{tab:chamferMannekin}
\end{table*}

\begin{figure}[bp]
\centering
    \includegraphics[width=0.45\textwidth]{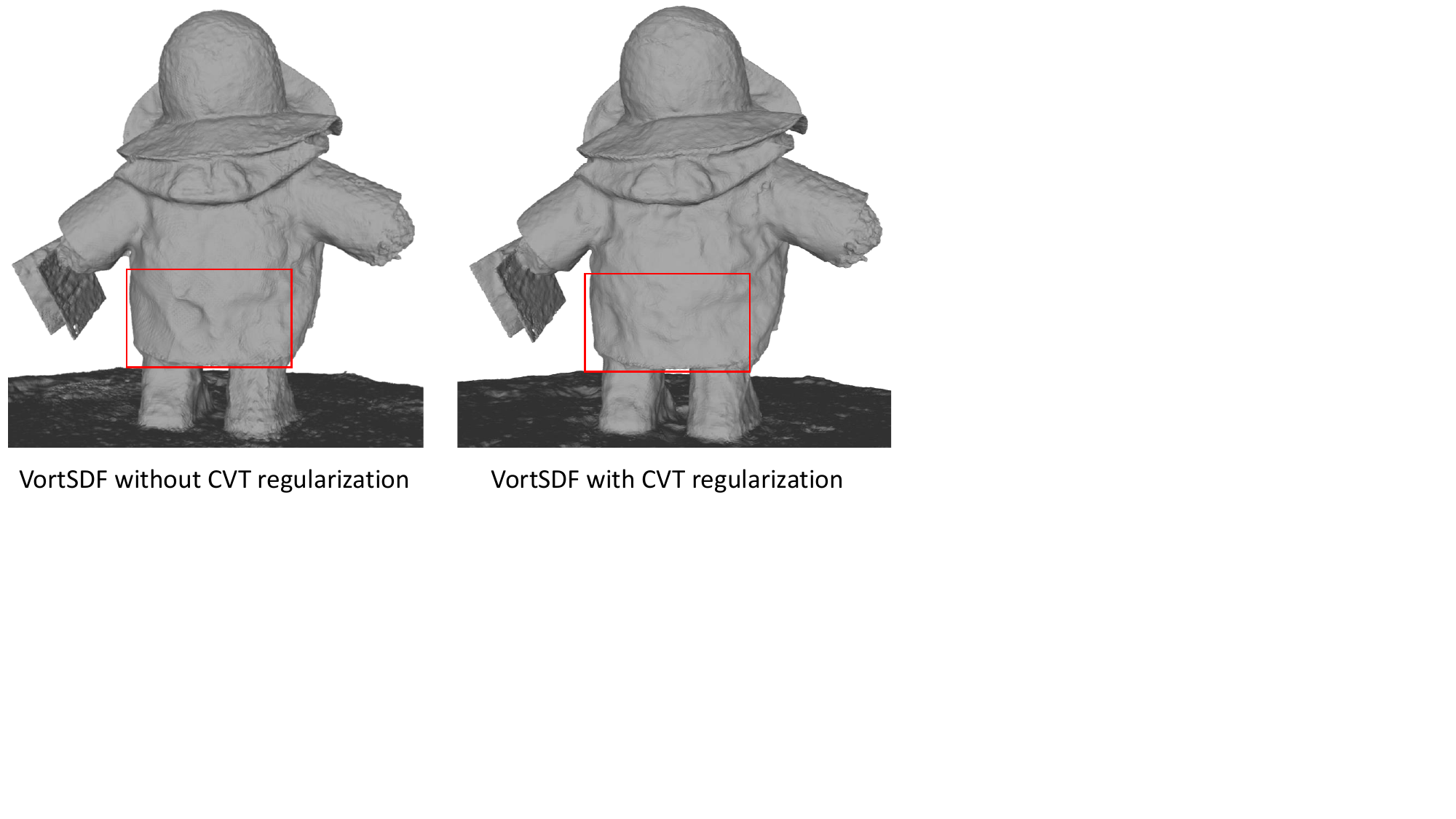}
    \caption{Regularizing the 3D discretization with our proposed approximated CVT loss significantly improve reconstruction quality.}
    \label{fig::resultsA}
\end{figure}

\subsection{Ablation study}
We evaluate the advantage of using our proposed CVT regularization on the Bear scene of Blended MVS. We run our proposed method with and without applying the CVT regularization after each up-sampling. Figure \ref{fig::resultsA} shows that regularizing the discretization around the surface significantly improves the reconstruction accuracy.

\subsection{Limitations}
Our proposed method has some limitations that can be addressed in future works. Mainly, at each level of up-sampling the tetrahedral mesh must be computed to do ray marching. We used an off-the-shelf software to compute the tetrahedral mesh (open3D) that is a CPU version of Delaunay triangulation and takes significant amount of time for millions of input points. Using more advanced GPU implementation of the Delaunay tetrahedralization construction algorithm would provide significant speedups.

%% file: main.bbl
\begin{thebibliography}{10}\itemsep=-1pt

\bibitem{aman2022compact}
Aytek Aman, Serkan Demirci, and U{\u{g}}ur G{\"u}d{\"u}kbay.
\newblock Compact tetrahedralization-based acceleration structures for ray
  tracing.
\newblock {\em Journal of Visualization}, 25(5):1103--1115, 2022.

\bibitem{4dhumanoutfit}
Matthieu Armando, Laurence Boissieux, Edmond Boyer, Jean-Sebastien Franco,
  Martin Humenberger, Christophe Legras, Vincent Leroy, Mathieu Marsot, Julien
  Pansiot, Sergi Pujades, Rim Rekik, Gregory Rogez, Anilkumar Swamy, and
  Stefanie Wuhrer.
\newblock 4dhumanoutfit: a multi-subject 4d dataset of human motion sequences
  in varying outfits exhibiting large displacements.
\newblock {\em Computer Vision and Image Understanding}.

\bibitem{barron2021mipnerf}
Jonathan~T Barron, Ben Mildenhall, Matthew Tancik, Peter Hedman, Ricardo
  Martin-Brualla, and Pratul~P Srinivasan.
\newblock Mip-nerf: A multiscale representation for anti-aliasing neural
  radiance fields.
\newblock In {\em Proceedings of the IEEE/CVF International Conference on
  Computer Vision (ICCV)}, pages 5855--5864, 2021.

\bibitem{barron2022mip}
Jonathan~T Barron, Ben Mildenhall, Dor Verbin, Pratul~P Srinivasan, and Peter
  Hedman.
\newblock Mip-nerf 360: Unbounded anti-aliased neural radiance fields.
\newblock In {\em Proceedings of the IEEE/CVF Conference on Computer Vision and
  Pattern Recognition (CVPR)}, pages 5470--5479, 2022.

\bibitem{barron2023zipnerf}
Jonathan~T. Barron, Ben Mildenhall, Dor Verbin, Pratul~P. Srinivasan, and Peter
  Hedman.
\newblock Zip-nerf: Anti-aliased grid-based neural radiance fields.
\newblock In {\em Proceedings of the IEEE/CVF International Conference on
  Computer Vision ({ICCV})}, 2023.

\bibitem{chen2022tensorf}
Anpei Chen, Zexiang Xu, Andreas Geiger, Jingyi Yu, and Hao Su.
\newblock Tensorf: Tensorial radiance fields.
\newblock In {\em European Conference on Computer Vision (ECCV)}, pages
  333--350. Springer, 2022.

\bibitem{chen2023factor}
Anpei Chen, Zexiang Xu, Xinyue Wei, Siyu Tang, Hao Su, and Andreas Geiger.
\newblock Factor fields: A unified framework for neural fields and beyond.
\newblock {\em arXiv preprint arXiv:2302.01226}, 2023.

\bibitem{chen2023neurbf}
Zhang Chen, Zhong Li, Liangchen Song, Lele Chen, Jingyi Yu, Junsong Yuan, and
  Yi Xu.
\newblock Neurbf: A neural fields representation with adaptive radial basis
  functions.
\newblock In {\em Proceedings of the IEEE/CVF International Conference on
  Computer Vision (ICCV)}, pages 4182--4194, 2023.

\bibitem{du99}
Qiang Du, Vance Faber, and Max Gunzburger.
\newblock Centroidal voronoi tessellations: applications and algorithms.
\newblock {\em SIAM review 41(4)}, 1999.

\bibitem{fridovich2023k}
Sara Fridovich-Keil, Giacomo Meanti, Frederik~Rahb{\ae}k Warburg, Benjamin
  Recht, and Angjoo Kanazawa.
\newblock K-planes: Explicit radiance fields in space, time, and appearance.
\newblock In {\em Proceedings of the IEEE/CVF Conference on Computer Vision and
  Pattern Recognition (CVPR)}, pages 12479--12488, 2023.

\bibitem{fridovich2022plenoxels}
Sara Fridovich-Keil, Alex Yu, Matthew Tancik, Qinhong Chen, Benjamin Recht, and
  Angjoo Kanazawa.
\newblock Plenoxels: Radiance fields without neural networks.
\newblock In {\em Proceedings of the IEEE/CVF Conference on Computer Vision and
  Pattern Recognition (CVPR)}, pages 5501--5510, 2022.

\bibitem{kellnhofer2021neural}
Petr Kellnhofer, Lars~C Jebe, Andrew Jones, Ryan Spicer, Kari Pulli, and Gordon
  Wetzstein.
\newblock Neural lumigraph rendering.
\newblock In {\em Proceedings of the IEEE/CVF Conference on Computer Vision and
  Pattern Recognition (CVPR)}, pages 4287--4297, 2021.

\bibitem{kerbl20233d}
Bernhard Kerbl, Georgios Kopanas, Thomas Leimk{\"u}hler, and George Drettakis.
\newblock 3d gaussian splatting for real-time radiance field rendering.
\newblock {\em ACM Transactions on Graphics (ToG)}, 42(4):1--14, 2023.

\bibitem{kulhanek2023tetra}
Jonas Kulhanek and Torsten Sattler.
\newblock Tetra-nerf: Representing neural radiance fields using tetrahedra.
\newblock {\em Proceedings of the IEEE/CVF International Conference on Computer
  Vision ({ICCV})}, 2023.

\bibitem{li2023neuralangelo}
Zhaoshuo Li, Thomas M\"uller, Alex Evans, Russell~H Taylor, Mathias Unberath,
  Ming-Yu Liu, and Chen-Hsuan Lin.
\newblock Neuralangelo: High-fidelity neural surface reconstruction.
\newblock In {\em IEEE Conference on Computer Vision and Pattern Recognition
  ({CVPR})}, 2023.

\bibitem{liu2020neural}
Lingjie Liu, Jiatao Gu, Kyaw~Zaw Lin, Tat-Seng Chua, and Christian Theobalt.
\newblock Neural sparse voxel fields.
\newblock {\em NeurIPS}, 2020.

\bibitem{liu2009centroidal}
Yang Liu, Wenping Wang, Bruno L{\'e}vy, Feng Sun, Dong-Ming Yan, Lin Lu, and
  Chenglei Yang.
\newblock On centroidal voronoi tessellation—energy smoothness and fast
  computation.
\newblock {\em ACM Transactions on Graphics (ToG)}, 28(4):1--17, 2009.

\bibitem{Lombardi2019TOG}
Stephen Lombardi, Tomas Simon, Jason Saragih, Gabriel Schwartz, Andreas
  Lehrmann, and Yaser Sheikh.
\newblock Neural volumes: Learning dynamic renderable volumes from images.
\newblock {\em ACM Trans. Graph.}, 38(4):65:1--65:14, July 2019.

\bibitem{lorensen1998marching}
William~E Lorensen and Harvey~E Cline.
\newblock Marching cubes: A high resolution 3d surface construction algorithm.
\newblock In {\em Seminal graphics: pioneering efforts that shaped the field},
  pages 347--353. 1998.

\bibitem{maruani2023voromesh}
Nissim Maruani, Roman Klokov, Maks Ovsjanikov, Pierre Alliez, and Mathieu
  Desbrun.
\newblock Voromesh: Learning watertight surface meshes with voronoi diagrams.
\newblock In {\em Proceedings of the IEEE/CVF International Conference on
  Computer Vision}, pages 14565--14574, 2023.

\bibitem{mildenhall2020nerf}
Ben Mildenhall, Pratul~P. Srinivasan, Matthew Tancik, Jonathan~T. Barron, Ravi
  Ramamoorthi, and Ren Ng.
\newblock Nerf: Representing scenes as neural radiance fields for view
  synthesis.
\newblock In {\em {European Conference on Computer Vision (ECCV)}}, 2020.

\bibitem{mueller2022instant}
Thomas M\"uller, Alex Evans, Christoph Schied, and Alexander Keller.
\newblock Instant neural graphics primitives with a multiresolution hash
  encoding.
\newblock {\em arXiv:2201.05989}, Jan. 2022.

\bibitem{oechsle2021unisurf}
Michael Oechsle, Songyou Peng, and Andreas Geiger.
\newblock Unisurf: Unifying neural implicit surfaces and radiance fields for
  multi-view reconstruction.
\newblock In {\em Proceedings of the IEEE/CVF International Conference on
  Computer Vision (ICCV)}, pages 5589--5599, 2021.

\bibitem{park2019deepsdf}
Jeong~Joon Park, Peter Florence, Julian Straub, Richard Newcombe, and Steven
  Lovegrove.
\newblock Deepsdf: Learning continuous signed distance functions for shape
  representation.
\newblock In {\em Proceedings of the IEEE/CVF conference on computer vision and
  pattern recognition (CVPR)}, pages 165--174, 2019.

\bibitem{ray2018meshless}
Nicolas Ray, Dmitry Sokolov, Sylvain Lefebvre, and Bruno L{\'e}vy.
\newblock Meshless voronoi on the gpu.
\newblock {\em ACM Transactions on Graphics (TOG)}, 37(6):1--12, 2018.

\bibitem{rebain2021derf}
Daniel Rebain, Wei Jiang, Soroosh Yazdani, Ke Li, Kwang~Moo Yi, and Andrea
  Tagliasacchi.
\newblock Derf: Decomposed radiance fields.
\newblock In {\em Proceedings of the IEEE/CVF Conference on Computer Vision and
  Pattern Recognition (CVPR)}, pages 14153--14161, 2021.

\bibitem{reiser2021kilonerf}
Christian Reiser, Songyou Peng, Yiyi Liao, and Andreas Geiger.
\newblock Kilonerf: Speeding up neural radiance fields with thousands of tiny
  mlps.
\newblock In {\em Proceedings of the IEEE/CVF International Conference on
  Computer Vision (ICCV)}, pages 14335--14345, 2021.

\bibitem{shen2021dmtet}
Tianchang Shen, Jun Gao, Kangxue Yin, Ming-Yu Liu, and Sanja Fidler.
\newblock Deep marching tetrahedra: a hybrid representation for high-resolution
  3d shape synthesis.
\newblock In {\em Advances in Neural Information Processing Systems (NeurIPS)},
  2021.

\bibitem{SunSC22}
Cheng Sun, Min Sun, and Hwann{-}Tzong Chen.
\newblock Direct voxel grid optimization: Super-fast convergence for radiance
  fields reconstruction.
\newblock In {\em IEEE Conference on Computer Vision and Pattern Recognition
  ({CVPR})}, 2022.

\bibitem{sun2022direct}
Cheng Sun, Min Sun, and Hwann-Tzong Chen.
\newblock Direct voxel grid optimization: Super-fast convergence for radiance
  fields reconstruction.
\newblock In {\em Proceedings of the IEEE/CVF Conference on Computer Vision and
  Pattern Recognition (CVPR)}, pages 5459--5469, 2022.

\bibitem{wang:hal-01185210}
Li Wang, Franck H{\'e}troy-Wheeler, and Edmond Boyer.
\newblock {A Hierarchical Approach for Regular Centroidal Voronoi
  Tessellations}.
\newblock {\em {Computer Graphics Forum}}, 35(1):152--165, Feb. 2016.

\bibitem{wang2016volumetric}
Li Wang, Franck H{\'e}troy-Wheeler, and Edmond Boyer.
\newblock On volumetric shape reconstruction from implicit forms.
\newblock In {\em European Conference on Computer Vision (ECCV)}, pages
  173--188. Springer, 2016.

\bibitem{wang2021neus}
Peng Wang, Lingjie Liu, Yuan Liu, Christian Theobalt, Taku Komura, and Wenping
  Wang.
\newblock Neus: Learning neural implicit surfaces by volume rendering for
  multi-view reconstruction.
\newblock {\em arXiv preprint arXiv:2106.10689}, 2021.

\bibitem{neus2}
Yiming Wang, Qin Han, Marc Habermann, Kostas Daniilidis, Christian Theobalt,
  and Lingjie Liu.
\newblock Neus2: Fast learning of neural implicit surfaces for multi-view
  reconstruction.
\newblock In {\em Proceedings of the IEEE/CVF International Conference on
  Computer Vision ({ICCV})}, 2023.

\bibitem{wu2022voxurf}
Tong Wu, Jiaqi Wang, Xingang Pan, Xudong Xu, Christian Theobalt, Ziwei Liu, and
  Dahua Lin.
\newblock Voxurf: Voxel-based efficient and accurate neural surface
  reconstruction.
\newblock In {\em International Conference on Learning Representations (ICLR)},
  2023.

\bibitem{yao2020blendedmvs}
Yao Yao, Zixin Luo, Shiwei Li, Jingyang Zhang, Yufan Ren, Lei Zhou, Tian Fang,
  and Long Quan.
\newblock Blendedmvs: A large-scale dataset for generalized multi-view stereo
  networks.
\newblock {\em Computer Vision and Pattern Recognition (CVPR)}, 2020.

\bibitem{yariv2020multiview}
Lior Yariv, Yoni Kasten, Dror Moran, Meirav Galun, Matan Atzmon, Basri Ronen,
  and Yaron Lipman.
\newblock Multiview neural surface reconstruction by disentangling geometry and
  appearance.
\newblock {\em Advances in Neural Information Processing Systems}, 33, 2020.

\bibitem{yu2021plenoctrees}
Alex Yu, Ruilong Li, Matthew Tancik, Hao Li, Ren Ng, and Angjoo Kanazawa.
\newblock Plenoctrees for real-time rendering of neural radiance fields.
\newblock In {\em Proceedings of the IEEE/CVF International Conference on
  Computer Vision (ICCV)}, pages 5752--5761, 2021.

\bibitem{kaizhang2020}
Kai Zhang, Gernot Riegler, Noah Snavely, and Vladlen Koltun.
\newblock Nerf++: Analyzing and improving neural radiance fields.
\newblock {\em arXiv:2010.07492}, 2020.

\end{thebibliography}
